\documentclass{IEEEtran}
\usepackage{amsmath,amsfonts}
\usepackage{graphicx}
\usepackage{cite}

\begin{document}

\title{Gradient Descent Algorithm Survey}
\author{DengFucheng, WangWanjie, GongAo, WangXiaoqi, WangFan}
\maketitle

\begin{abstract}
Focusing on the practical configuration needs of optimization algorithms in deep learning, this article concentrates on five major algorithms: SGD, Mini-batch SGD, Momentum, Adam, and Lion. It systematically analyzes the core advantages, limitations, and key practical recommendations of each algorithm. The research aims to gain an in-depth understanding of these algorithms and provide a standardized reference for the reasonable selection, parameter tuning, and performance improvement of optimization algorithms in both academic research and engineering practice, helping to solve optimization challenges in different scales of models and various training scenarios.
\end{abstract}


\section{Introduction}

\IEEEPARstart{O}{ptimization} algorithms are the backbone of deep-learning training; their design directly determines how fast a model converges, how stable the training process is, and how well the final model generalizes.  Starting from plain gradient descent, the community has introduced a long lineage of variants to cope with ever-larger models and increasingly diverse tasks.  As parameter counts have grown exponentially, classical batch gradient descent has revealed critical weaknesses—prohibitive computational cost, slow convergence, and severe oscillations in high-curvature regions—spurring the development of ideas based on stochastic sampling, momentum accumulation, adaptive learning-rate scaling, and even sign-based updates.

A systematic comparison of these gradient-descent variants not only deepens our theoretical understanding of training dynamics, but also gives practitioners a fast track to choosing the right optimizer.  To this end, we conduct a comprehensive study of five representative algorithms:SGD – the canonical stochastic method, valued for its solid theoretical grounding and strong generalization;Mini-batch SGD – the de-facto industry standard that balances computational efficiency with variance control;Momentum – accelerates convergence and damps oscillations by accumulating a velocity vector along the optimization trajectory;Adam – combines the best of RMSProp and Momentum, offering per-parameter adaptive learning rates;Lion – a recent, sign-based, momentum-style optimizer whose minimalistic update rule has proven both memory-efficient and remarkably robust on large-scale models.

Through principled derivations, optimization-path visualizations, and extensive empirical tests across multiple datasets, we evaluate these algorithms on convergence speed, stability, generalization performance, and compute/memory overhead.  The result is a concise, evidence-backed guide for selecting the most suitable optimizer under varying task requirements, model sizes, and hardware budgets.

\section{A Survey of Gradient-Descent Algorithms}

\subsection{SGD}

\subsubsection{Historical Context}

Stochastic gradient descent is the cornerstone algorithm for large-scale optimization and machine learning.  Viewed abstractly, it is the direct translation of classical stochastic-approximation methods to optimization problems.  Its theoretical roots trace back to the 1951 Robbins–Monro framework, which first formalized recursive root-finding under noisy observations and thereby laid the mathematical bedrock for all later sample-based optimizers.  Polyak and others subsequently studied trajectory averaging and related techniques, dramatically improving convergence speed and variance properties.  More recently, Bottou, Curtis, and Nocedal provided a unified treatment of SGD and its most common variants, bridging modern practice and theory \cite{ref1}.

\subsubsection{Algorithmic Description and Key Ideas}

\paragraph{Problem formulation}

We wish to minimize either expected or empirical risk  

\begin{equation}
\min_{w\in\mathbb{R}^d} F(w) = \mathbb{E}_{\xi}[f(w;\xi)] \quad\text{or}\quad F_n(w)=\frac{1}{n}\sum_{i=1}^n f(w;x_i), 
\end{equation}

where $f(w;\xi)$ is a differentiable loss (e.g., negative log-likelihood or squared error) and $\xi$ denotes a random data instance or mini-batch.

\paragraph{Update rule}

At iteration $t$ draw $\xi_t$ and perform  

\begin{equation}
w_{t+1}=w_t - \eta_t\, \nabla f(w_t;\xi_t).
\end{equation}

\paragraph{Key statistical properties}

1.Unbiasedness: If $\xi_t$ are i.i.d., then $\mathbb{E}[\nabla f(w;\xi)]=\nabla F(w)$, so the stochastic gradient is, on average, the true descent direction \cite{ref2}.2.Variance and oscillation: Single-sample gradients have large variance, causing the iterate sequence to fluctuate around the optimum.  Variance can be reduced by larger batch sizes, iterate averaging, or specialized variance-reduction techniques \cite{ref2}.

\subsubsection{Theoretical Highlights}

\paragraph{Learning-rate schedules and basic convergence}

Convergence hinges on the learning-rate sequence $\{\eta_t\}$ and on the curvature of $F$.  Typical conditions are $\sum_t \eta_t = \infty$ and $\sum_t \eta_t^2 < \infty$ (e.g., $\eta_t=\eta_0/(1+\lambda t)$).  Under smoothness and bounded-second-moment assumptions, SGD converges almost surely to a stationary point (non-convex case) or to the global minimizer (convex/strongly-convex case) \cite{ref1}.  
*Rate*: For convex Lipschitz losses the expected optimality gap is $O(1/\sqrt{T})$ after $T$ iterations; for strongly-convex losses an $O(1/T)$ rate is achievable with an appropriately decaying schedule \cite{ref2}.

\paragraph{Polyak averaging}

Polyak and Juditsky’s iterate averaging reduces the leading constant in mean-squared error to the statistical optimum, recovers asymptotic normality under mild conditions, and improves finite-sample behavior.  The technique is both theoretically optimal and practically effective for variance control \cite{ref3}.

\subsubsection{Summary}

SGD strikes a powerful computational–statistical balance that has made it the workhorse of large-scale learning.  Its simple update, linear scalability with sample size, and compatibility with momentum, mini-batching, and learning-rate heuristics keep it dominant in both industry and academia.  Current research continues to refine convergence rates, variance characterizations, and averaging schemes, while engineering efforts focus on hardware-aligned and distributed variants.

\subsection{Mini-Batch Stochastic Gradient Descent}

\subsubsection{Background and Development}

Batch Gradient Descent (BGD) requires computing the gradient using the entire training dataset at each iteration. As dataset sizes expand to millions or even larger scales, the computational cost of a single iteration becomes extremely high, making it unsuitable for large-scale learning tasks. The convergence of SGD was proven by Robbins and Monro through the stochastic approximation method \cite{ref1}. SGD uses one sample to update the gradient at each step, resulting in low computational cost but high gradient variance and unstable updates. The mini-batch strategy has gradually become the mainstream in practice, especially with the rise of large-scale machine learning and deep learning. Bottou emphasized the practical value of mini-batches in his research on large-scale learning \cite{ref5}, while systematic monographs and reviews on deep learning have further standardized this approach \cite{ref6, ref7}. Mini-batch SGD achieves an optimal balance between stability, high-frequency updates, and GPU parallel acceleration \cite{ref2}.

\subsubsection{Core Principles}

The goal of Mini-batch SGD remains to minimize the empirical risk:

\begin{equation}
\min_\theta L(\theta)=\frac{1}{N}\sum_{i=1}^N \ell(f(x_i;\theta), y_i)
\end{equation}

where: $N$ denotes the number of training samples, $\ell$ is the loss function, and $\theta$ represents the model parameters.

\paragraph{Mini-Batch Gradient Estimation}

At each iteration, a random batch is selected:

\begin{equation}
B_t = \{(x_i, y_i)\}_{i=1}^b
\end{equation}

The gradient estimate is:

\begin{equation}
g_t = \frac{1}{b}\sum_{i\in B_t} \nabla_\theta \ell(f(x_i;\theta_t), y_i)
\end{equation}

This estimation combines the efficiency of SGD and the stability of BGD \cite{ref2, ref5}.

\paragraph{Parameter Update}

\begin{equation}
\theta_{t+1} = \theta_t - \eta g_t
\end{equation}

where $\eta$ is the learning rate.

\paragraph{Noise Characteristics of Mini-Batches}

The noise introduced by mini-batches has a "regularization" effect, which is believed to help deep learning models escape local minima and saddle points \cite{ref6, ref7}. Ruder's review of optimization algorithms also highlights its impact on convergence and generalization \cite{ref8}.

\subsubsection{Advantages and Disadvantages}

\paragraph{Advantages}

Mini-batch SGD offers multiple core advantages: it is inherently compatible with the parallel computing capabilities of GPUs, providing a hardware-adapted foundation for improving training efficiency \cite{ref6}; compared to SGD, the algorithm exhibits smaller gradient variance and more stable training processes \cite{ref3}; meanwhile, the moderate noise contained in the gradient can effectively prevent model overfitting and help enhance generalization ability; these characteristics make it highly applicable in large-scale data training scenarios \cite{ref5}.

\paragraph{Disadvantages}

Despite the aforementioned advantages, Mini-batch SGD still has certain limitations: there is no universally optimal configuration for batch size, which requires careful parameter tuning based on specific tasks; excessively large batches tend to degrade model convergence and make it difficult to achieve ideal performance; at the same time, the algorithm may still be affected by saddle points, thereby hindering the efficient progress of the training process; in addition, the setting of the learning rate needs to be combined with auxiliary strategies such as warmup and decay to effectively avoid training fluctuations and ensure convergence stability and final performance.

\subsubsection{Summary}

Mini-batch SGD is a core algorithm for deep learning training. It combines the advantages of BGD and SGD, maintaining high computational efficiency while achieving good convergence characteristics. With the continuous development of hardware architectures, Mini-batch SGD remains one of the most practically valuable optimizers currently available and will long serve as a fundamental tool in deep learning.

\subsection{Momentum}

\subsubsection{Background and Development}

Polyak was the first to systematically propose the "momentum" idea in 1964, introducing the so-called heavy-ball method that incorporates an inertial term into iterative algorithms to accelerate convergence. From the perspective of numerical analysis and iterative methods, this work presented the form of multi-step iteration and several convergence results \cite{ref9}.

With the rapid development of machine learning and neural networks, the momentum method has been widely integrated into stochastic optimization frameworks and proven to be one of the key techniques in deep network training. Sutskever et al. systematically studied the importance of initialization and momentum for deep model training, emphasizing the significant impact of momentum on practical performance \cite{ref10}.

Subsequently, the optimization community further linked momentum with more powerful acceleration methods (such as Nesterov Accelerated Gradient) and investigated the dynamic behavior of momentum methods from the perspective of continuous limits (ordinary differential equations), providing new tools for understanding oscillations, overshoots, and convergence rates \cite{ref11}.

\subsubsection{Algorithm Formulations}

There are two commonly used discrete iterative formulations of the momentum method:

\paragraph{Polyak (Heavy-Ball) Formulation}
\begin{equation} 
v_{t+1} = \beta v_t - \eta \nabla f(\theta_t),\qquad  
\theta_{t+1} = \theta_t + v_{t+1},   
\end{equation}
where $\theta_t$ denotes the parameters, $v_t$ is the velocity (cumulative momentum), $\eta>0$ is the step size (learning rate), and $\beta \in [0,1)$ is the momentum coefficient (inertial factor). This formulation corresponds to Polyak's heavy-ball update \cite{ref9}.

\paragraph{Common Engineering/Deep Learning Implementation (With Weighted Historical Gradients)}

\begin{equation}
m_{t+1} = \beta m_t + (1-\beta)\nabla f(\theta_t),\qquad  
\theta_{t+1} = \theta_t - \eta m_{t+1}.  
\end{equation}

When $(1-\beta)$ is omitted or rewritten as 1 (i.e., $m$ stores an unscaled exponential moving average), the two formulations are equivalent to the scaling transformation of the previous form in practice. Deep learning frameworks (e.g., TensorFlow, PyTorch) typically adopt the first or the second equivalent implementation.

\subsubsection{Principles}

\paragraph{Intuitive Understanding}

The momentum method can be understood through the physical analogy of "mass-damping": the optimization path is regarded as the motion of a particle on the potential energy surface $f(\theta)$, where momentum $v$ is analogous to velocity (inertia), coefficient $\beta$ to the damping factor, and learning rate $\eta$ to the time scale and force amplification/reduction. Thus, momentum accumulates updates along consistent directions (acceleration), while offsetting partial noise when the gradient direction changes rapidly, introducing a smoothing effect that reduces local oscillations and improves convergence speed \cite{ref9, ref11}.

\paragraph{Second-Order Dynamical Perspective}

In the limit of small step sizes, the discrete update with momentum can be approximated by a second-order ordinary differential equation (ODE), for example:  
\begin{equation}
\ddot{\theta}(t) + c\dot{\theta}(t) + \nabla f(\theta(t)) = 0,  
\end{equation}
where $\ddot{\theta}$ represents acceleration, $\dot{\theta}$ denotes velocity, and $c$ is the damping coefficient. This ODE describes a damped Newtonian mechanical system. Researchers can analyze momentum-induced oscillations, damping, and convergence behavior through this ODE, explaining why appropriate damping (i.e., $\beta$ should not be excessively large) can prevent long-term oscillations \cite{ref11}.

\paragraph{Impact on SGD}

In stochastic gradient scenarios, momentum accumulates signals with stable directions on the one hand and filters out high-frequency noise on the other. This means that SGD with momentum can often escape saddle points faster and reach better generalization points in deep learning training. However, it may also cause overshoot when crossing narrow valleys due to inertia. Therefore, in practice, it is necessary to coordinately tune the learning rate, momentum coefficient, step size decay, and batch size \cite{ref9, ref10}.

\subsubsection{Theoretical Convergence}

For convex and smooth objective functions, Polyak's heavy-ball method can achieve accelerated linear convergence under certain strongly convex conditions. Nesterov's accelerated method provides an explicit $O(1/k^2)$ non-asymptotic optimal convergence rate in convex optimization (compared to $O(1/k)$ for GD), while the convergence rate analysis of Polyak-heavy-ball depends on damping and spectral characteristics \cite{ref9, ref11}.

In stochastic optimization scenarios, the strict convergence properties of SGD with momentum require more refined noise models and step size scheduling assumptions. In practice, fixing the momentum coefficient and using appropriate learning rate decay can often ensure good performance, but the theoretical global convergence for non-convex deep networks remains an open problem \cite{ref10, ref11}.

\subsubsection{Engineering Practice}

In the practical application of the momentum method, the following configuration guidelines verified by academic research and practice can be followed: the momentum coefficient $\beta$ is commonly set to 0.9 or 0.99. For RNN or deep network training, Sutskever et al. suggested achieving more robust training effects through the joint design of momentum parameters and initialization strategies \cite{ref10}; regarding the learning rate $\eta$, the momentum mechanism allows the use of a slightly larger learning rate than pure SGD to accelerate convergence, but excessive values should be avoided to prevent training divergence caused by the superposition of momentum inertia, and it is usually necessary to use it with learning rate decay strategies; the choice of batch size needs to balance gradient noise and generalization ability—larger batches can reduce gradient noise, making the effective signals accumulated by momentum more reliable, but excessively increasing the batch size may lead to a decline in model generalization performance; in addition, the concept of momentum has been integrated into adaptive optimizers. Essentially, these methods implement exponential weighted averaging of gradients, but the optimization behavior brought by adaptive step sizes differs from that of pure momentum methods. In practical use, it is necessary to trade off generalization performance and convergence speed according to task requirements \cite{ref10}.

\subsubsection{Summary}

By introducing the accumulation of historical gradients, the momentum method provides a simple yet effective acceleration mechanism, which has almost become a standard component in deep learning training. Its dynamics can be described by second-order ODEs, offering theoretical tools for explaining oscillations and overshoots. Although the rigorous global convergence for non-convex deep networks remains to be further studied, from an engineering perspective, reasonable configuration of the momentum coefficient and learning rate, combined with modern scheduling strategies and lookahead variants, can significantly improve training efficiency and robustness.

\subsection{Adam}

\subsubsection{Background and Development}

Adaptive learning rate methods improve convergence efficiency and robustness by adjusting the step size of each parameter coordinate based on historical gradients. Proposed by Kingma and Ba between 2014 and 2015, Adam combines first-order momentum with second-order momentum, and introduces bias correction to mitigate estimation bias in the initial iteration stage. This results in a general-purpose optimizer that performs well on problems with sparse gradients or high noise. The proposal and experimental validation of Adam laid the foundation for its widespread adoption in deep learning practice, and it remains one of the most commonly used adaptive first-order optimizers in deep learning training to date \cite{ref12}.

\subsubsection{Core Algorithm and Mathematical Principles}

\paragraph{Optimization Objective and Notation}

Let the objective function be the empirical risk  
\begin{equation}
F(w)=\frac{1}{n}\sum_{i=1}^n f(w;x_i),  
\end{equation} or the more general expected form $F(w)=\mathbb{E}_\xi[f(w;\xi)]$. At iteration $t$, the stochastic gradient $g_t=\nabla f(w_t;\xi_t)$ is computed.

\paragraph{Key Quantities of Adam (Momentum and Second-Order Moment Estimation)}

Adam uses exponential moving averages (EMA) to estimate the first moment (mean) and second moment (uncentered variance approximation) of the gradients, respectively:  
\begin{equation}
m_t = \beta_1 m_{t-1} + (1-\beta_1) g_t,\qquad  
v_t = \beta_2 v_{t-1} + (1-\beta_2) g_t^2,  
\end{equation} where the operations are element-wise, and $\beta_1,\beta_2\in[0,1)$ are decay factors.

Bias Correction: Since $m_0=v_0=0$, the original EMA has a bias toward zero in the early iterations. Kingma and Ba proposed bias correction as follows:  
\begin{equation}
\hat m_t = \frac{m_t}{1-\beta_1^t},\qquad  
\hat v_t = \frac{v_t}{1-\beta_2^t}.  
\end{equation} After correction, the estimators are closer to the true moments in the early iterations \cite{ref12}.

\paragraph{Step Size Rule (Element-Wise Normalization)}

Adam adopts element-wise scaled updates:  
\begin{equation}
w_{t+1} = w_t - \eta \frac{\hat m_t}{\sqrt{\hat v_t} + \epsilon},  
\end{equation}
where $\eta$ is the global step size (learning rate), and $\epsilon$ is a numerical stability term (typically set to $10^{-8}$). Since the denominator is the square root of the second moment, coordinates with larger gradient magnitudes will have smaller step sizes, while those with smaller magnitudes will have larger step sizes—this gives the algorithm a certain degree of invariance to parameters of different scales \cite{ref12}.

\subsubsection{Engineering Practice}

In the practical application of the Adam algorithm, the following common configurations can be referenced: For hyperparameter settings, the default combination $\beta_1=0.9, \beta_2=0.999, \epsilon=10^{-8}$ serves as a reliable starting point for most vision and language model training. The learning rate $\eta$ needs to be flexibly adjusted based on specific tasks, and is usually combined with learning rate warmup and decay strategies to optimize the convergence process \cite{ref12}; For weight decay strategies, for networks requiring regularization, it is recommended to prioritize the AdamW variant with decoupled weight decay over directly adding an L2 regularization term to the loss function. Especially in performance comparison experiments between Adam and SGD, this choice better ensures the fairness of the comparison \cite{ref14}; For the coordination between batch size adjustment and learning rate scaling, when the batch size is significantly increased, the learning rate should be adjusted upward accordingly, and the warmup phase should be extended to mitigate instability in the early training stage \cite{ref12}; For generalization performance optimization and model validation, if the goal is to pursue optimal test performance, SGD (+ momentum) is recommended as the baseline comparison algorithm. In the actual development process, Adam can first be used to achieve rapid convergence, complete model prototyping and hyperparameter search, and then switch to SGD for fine-tuning as needed. If Adam is continuously used, trying the AdamW variant with a reasonable learning rate decay strategy can often further improve the model's generalization ability \cite{ref14}.

\subsubsection{Conclusion}

Adam cleverly combines momentum with adaptive second-order moment estimation, providing a practical solution to problems with noisy and inconsistent gradient scales. With its fast convergence and minimal hyperparameter tuning requirements, it has become a common baseline for deep learning training. Meanwhile, theoretical studies have shown that Adam may have issues with convergence and generalization discrepancies in certain scenarios; subsequent works (such as AdamW) have corrected or improved the original algorithm from various perspectives \cite{ref12}.

\subsection{Lion}

\subsubsection{Background and Development}

Lion is a first-order optimizer automatically discovered through program/symbolic search, centered on symbolic "momentum-augmented sign updates". Its core idea is: first mix the current gradient with historical gradients using momentum, then take the sign of the mixed result, and update the parameters with a fixed magnitude. This design makes Lion superior to the Adam series in memory usage (only needing to maintain momentum instead of matrix second-order moment estimation) and has demonstrated superior or comparable performance on several large-scale pre-training/fine-tuning tasks for vision, language, and generative models \cite{ref15, ref16}.

Lion originates from a "symbolic algorithm discovery" work by Google, which formalizes "optimizer discovery" as a program search problem. Efficient optimizers are discovered from an infinite space through evolution/search and evaluation on proxy tasks, ultimately leading to the proposal of Lion. Its design integrates the idea of momentum with "symbolic" updates, but unlike adaptive methods, it does not explicitly maintain second-order moment estimation. Instead, it uses the sign operation to make the update magnitude of each parameter the same. This feature brings simplicity in memory and implementation, and improves efficiency in certain large-scale training scenarios.

\subsubsection{Algorithm Formulation}

\paragraph{Notation and Variables}

Let $\theta_t\in\mathbb{R}^d$ be the parameter vector, $g_t = \nabla_\theta \ell(\theta_t)$ be the (mini-batch) gradient. Hyperparameters include the learning rate $\eta>0$, two momentum coefficients $\beta_1, \beta_2\in[0,1)$ (commonly $\beta_1=0.9, \beta_2=0.99$ or similar values in the paper), and an optional decoupled weight decay coefficient $\lambda$.

\paragraph{Analytical Formulas}

Vector formula form (omitting the weight decay term):

\begin{equation}
\begin{aligned} 
m_{t} &= \beta_1 m_{t-1} + (1-\beta_1) g_t,\\ 
u_{t} &= \beta_2 m_{t-1} + (1-\beta_2) g_t,\\ 
\theta_{t+1} &= \theta_t - \eta \, \operatorname{sign}(u_t). 
\end{aligned}
\end{equation}

where $\operatorname{sign}(\cdot)$ takes the sign of each component of the vector. Note: There are slight engineering variants in the interpolation coefficients and order of $m$ and $u$ in the paper; the above is the most common and easily understandable form. During implementation, consistency in details (such as whether $(1-\beta)$ is included in the EMA) should be maintained to reproduce results \cite{ref15}.

\subsubsection{Principles}

\paragraph{Role of Symbolic Updates (sign Operation)}

First, unify the update magnitude: taking the sign of $u_t$ means that the update magnitude of all parameter components at each step is controlled by the same scalar $\eta$ (the direction is determined by the sign). This is opposite to adaptive methods such as Adam, which scale the update magnitude component-wise, resulting in more uniform step size behavior \cite{ref15}. Second, the sign operation reduces sensitivity to gradient magnitude, which can act as a filtering effect under high-noise mini-batch gradients (retaining only direction information), thereby enabling more stable long-step training on certain tasks. However, it may also lose important information about the magnitude \cite{ref15}.

\paragraph{Momentum and Interpolation}

The paper uses two sets of interpolations, essentially constructing the input $u_t$ for the sign operation using memory-augmented mixed signals, thereby balancing historical gradient directions with the current gradient. Compared to directly taking the sign of the current gradient, the accumulation of historical information from interpolation makes the optimization path smoother against "short-term noise" while still accumulating effective signals in the long-term direction \cite{ref15}.

\subsubsection{Advantages, Limitations, and Application Recommendations}

\paragraph{Advantages}

The Lion algorithm has three core advantages: First, excellent memory efficiency—only first-order momentum needs to be maintained (instead of the second-order moment estimation required by Adam), which can significantly reduce memory overhead in ultra-large model training scenarios \cite{ref15}; Second, strong adaptability to large-scale tasks—empirically validated to achieve performance improvements in pre-training and fine-tuning tasks for various large models such as vision and generative models \cite{ref15}; Third, low implementation threshold—the algorithm logic is concise and intuitive, enabling convenient deployment and integration in mainstream deep learning frameworks.

\paragraph{Limitations and Risks}

Despite the above advantages, the Lion algorithm still has several limitations that need improvement: First, it is sensitive to the learning rate—smaller learning rates than those recommended for Adam are required, combined with reasonable learning rate scheduling strategies to avoid instability in early training \cite{ref15}; Second, its performance advantages are scenario-dependent—empirical results in the paper show that Lion's optimization effect gradually improves with increasing batch size. In small-batch training or small-scale model tasks, its performance may not be superior to traditional optimization algorithms such as Adam; Third, relatively weak theoretical support—compared to the rigorous convergence theory systems formed by algorithms such as SGD and Adam under convex or weakly convex optimization settings, Lion's design based on symbolic updates and automatic algorithm discovery still has many unclear gaps in theoretical analysis. Related theoretical verification and improvement work require further research \cite{ref15}.

\paragraph{Practical Recommendations}

In the practical application of the Lion algorithm, the following optimized configuration recommendations verified by the paper and community practice can be referenced: For hyperparameter settings, the default recommendation is $\beta_1\approx0.9$ and $\beta_2\approx0.99$. The learning rate needs to be set to a smaller value than the recommended value for Adam, and the optimal specific value is recommended to be determined through grid search in small-scale pre-experiments \cite{ref15}; For training strategies, introducing a learning rate warmup process of several thousand steps can effectively mitigate training instability in the initial stage, which can be followed by cosine decay or multi-stage step decay strategies to further optimize convergence; For weight decay mechanisms, decoupled weight decay is recommended, which is more compatible with Lion's update logic and improves the reproducibility of experimental results in the paper; For scenario adaptation, if large-batch training capabilities or distributed training resources are available, prioritizing the Lion algorithm can more fully leverage its performance advantages \cite{ref15}.

\subsubsection{Summary}

Lion is a novel first-order optimizer discovered through automatic search from numerous candidate algorithms, centered on "momentum-augmented sign updates". It shows significant potential in memory savings and performance on large-scale training tasks, but is sensitive to learning rate and batch size, and its theoretical properties are still under research. For engineering teams looking to explore alternatives to Adam in ultra-large model/large-batch scenarios, Lion is a powerful candidate worth trying; in small-model or resource-constrained settings, careful comparison and simultaneous hyperparameter tuning should be conducted.

\section{Experimental Comparison and Analysis}

This section conducts a systematic experimental comparison of five typical gradient descent optimization algorithms based on the MNIST handwritten digit classification dataset and the California Housing regression dataset.

\subsection{Experimental Setup}

In terms of experimental data, MNIST is used to evaluate performance on classification tasks, with inputs as 28×28 grayscale images and outputs as ten-class digit labels; the California Housing dataset constitutes a typical structured regression scenario, where the prediction target is the logarithmic mean of housing prices, and the standard evaluation metric is Root Mean Squared Error (RMSE). To focus the comparison on the inherent performance of the optimizers, all models adopt a concise multi-layer perceptron (MLP) structure with the same number of layers and activation functions. The MNIST model consists of two fully connected layers with ReLU activation, while the housing price regression model incorporates additional hidden layers and Dropout to enhance generalization. During training, all optimization algorithms use the same learning rate, batch size, and number of training epochs. The additional hyperparameters of the optimizers themselves adopt the standard settings recommended in the literature to minimize the interference of hyperparameter tuning factors. This design ensures that the experimental results can accurately reflect the inherent differences between optimization strategies.

\subsection{Convergence Performance Comparison}

\begin{figure}[htbp]
\centering
\includegraphics[width=2.5in]{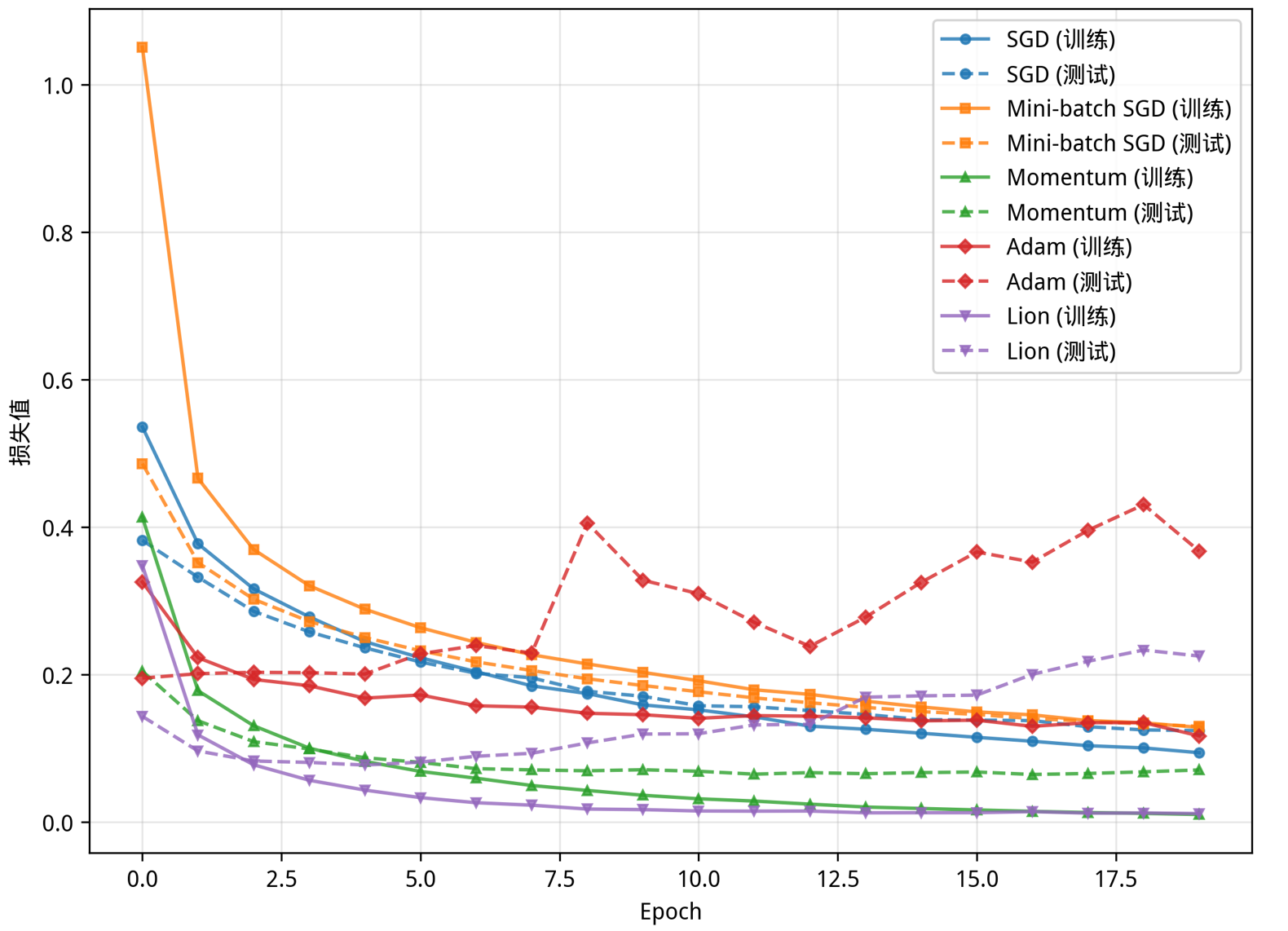}
\caption{Loss Curves of Different Algorithms on MNIST}
\label{fig_1}
\end{figure}

\begin{figure}[htbp]
\centering
\includegraphics[width=2.5in]{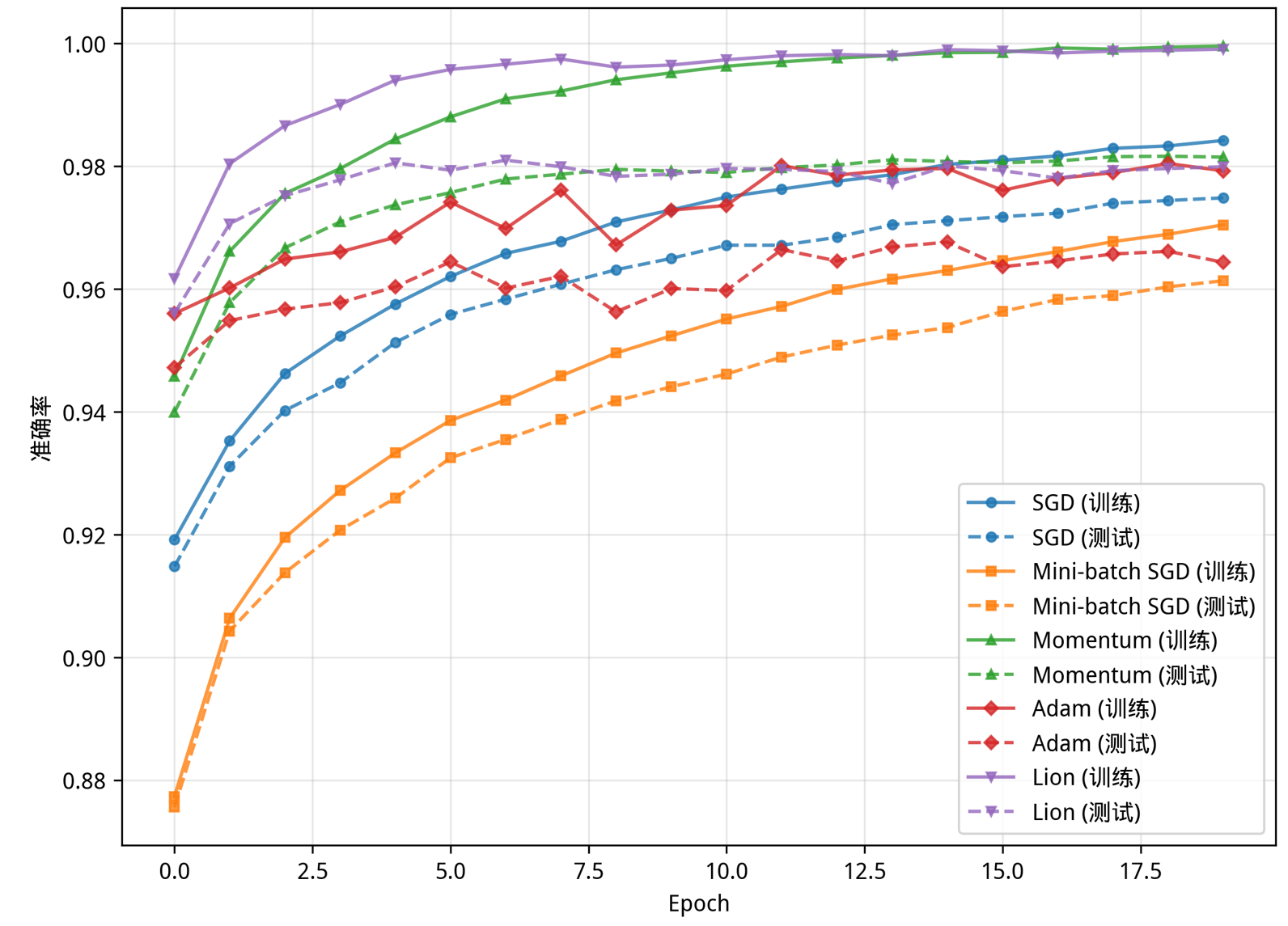}
\caption{Performance Curves of Different Algorithms on MNIST}
\label{fig_2}
\end{figure}

From the training loss curves on MNIST, it can be observed that Momentum and Lion are significantly leading in convergence speed. They achieve low loss values within the first few epochs and maintain a steady decline in the later stages. In contrast, although SGD has a slower descent rate, its curve is very smooth, demonstrating good stability. This is related to its update mechanism—even without a momentum term, its process of gradually approaching the optimal solution remains reliable. Mini-batch SGD performs the worst initially, with significantly higher loss and obvious fluctuations, indicating that its noise level is relatively large, making it difficult to quickly determine the descent direction. Adam declines rapidly in the early stage but exhibits obvious jitter in the middle stage, which is associated with the sensitivity of its adaptive method to noise distribution under default hyperparameters.

In terms of test accuracy on MNIST, Momentum and Lion achieve the optimal accuracy of approximately 0.98, overall leading other algorithms. Although the final accuracy of SGD is slightly lower, it gradually approaches the optimal value with training, maintaining good generalization ability. Both Adam and Mini-batch SGD experience varying degrees of performance degradation, indicating that they fail to fully stabilize the gradient update process under the current hyperparameter settings.

\begin{figure}[htbp]
\centering
\includegraphics[width=2.5in]{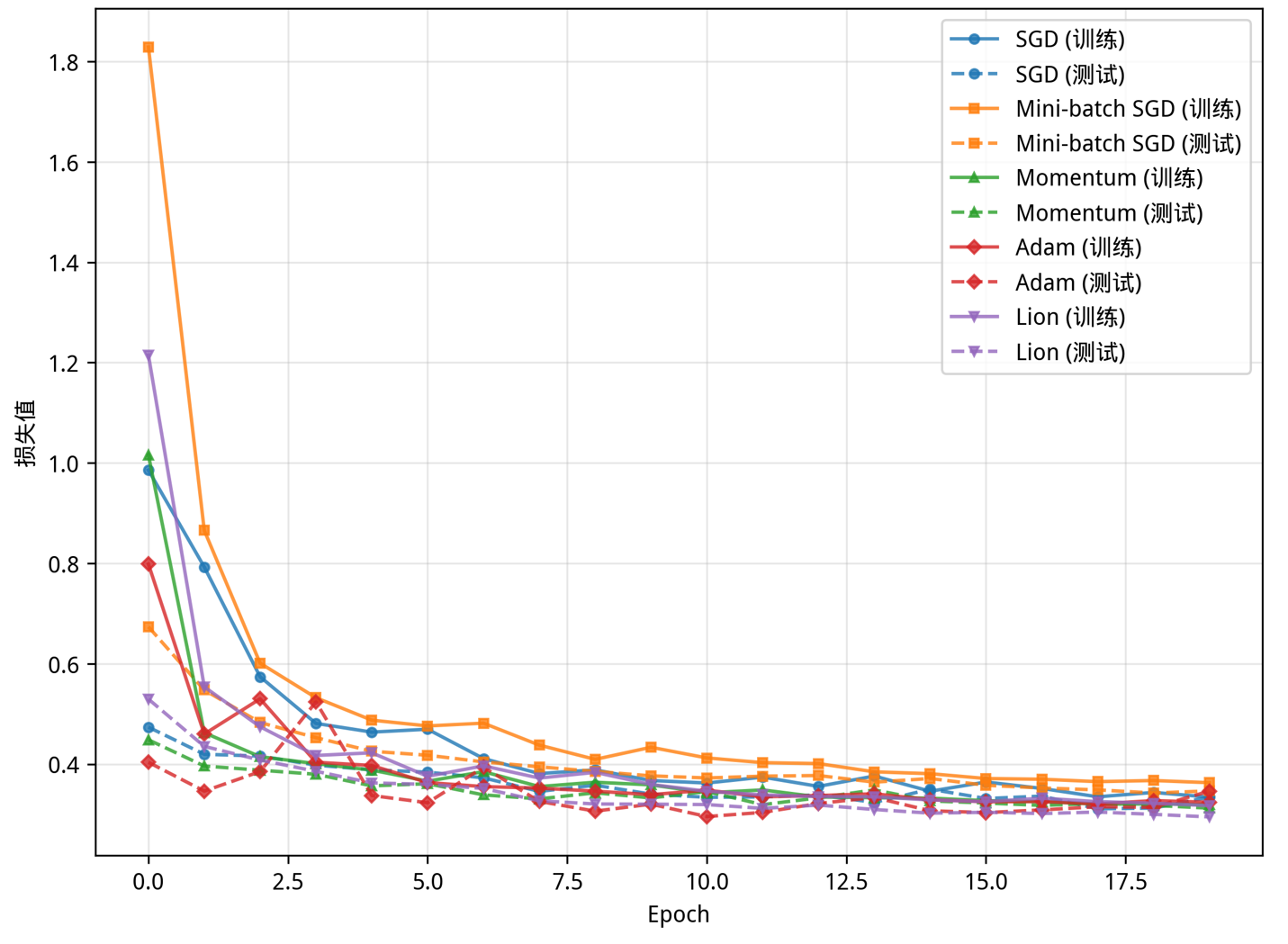}
\caption{Loss Curves of Different Algorithms on California Housing}
\label{fig_3}
\end{figure}

\begin{figure}[htbp]
\centering
\includegraphics[width=2.5in]{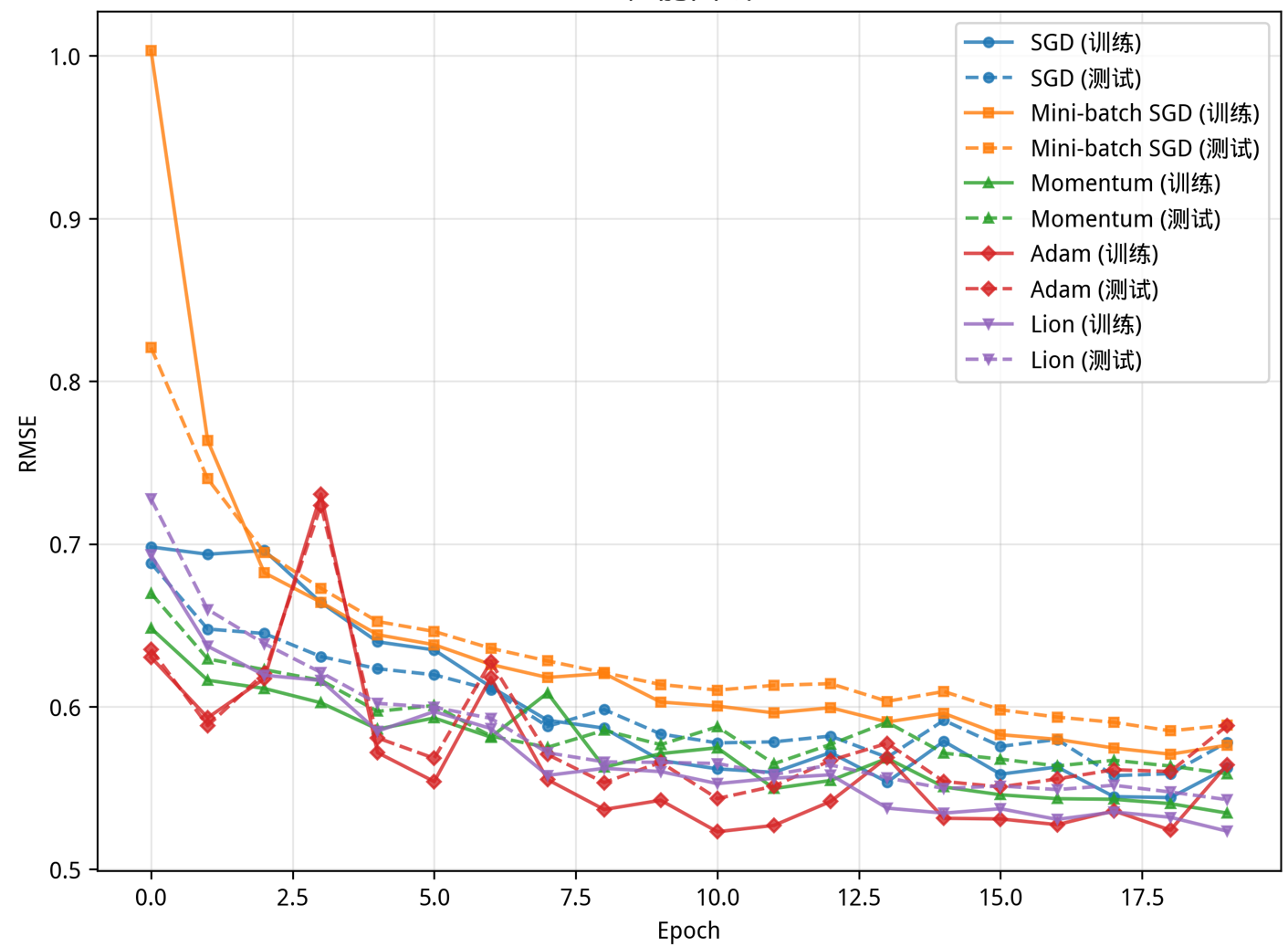}
\caption{Performance Curves of Different Algorithms on California Housing}
\label{fig_4}
\end{figure}

In the California Housing task, the convergence speed of all optimizers is much slower than that on MNIST. This is a natural phenomenon caused by the higher noise and more complex objective function surface of regression tasks. On this task, Lion performs the most prominently, with its RMSE curve declining rapidly and having the smallest fluctuation amplitude, reflecting the advantage of its gradient sign update strategy on noisy data. Momentum still demonstrates good stability, while SGD performs moderately with a slow but stable descent. Both Mini-batch SGD and Adam exhibit relatively unstable performance in the regression task, with multiple spikes on their loss curves, indicating that they are more sensitive to feature noise and changes in gradient scales.

\subsection{Gradient Characteristic Comparison}

\begin{figure}[htbp]
\centering
\includegraphics[width=2.5in]{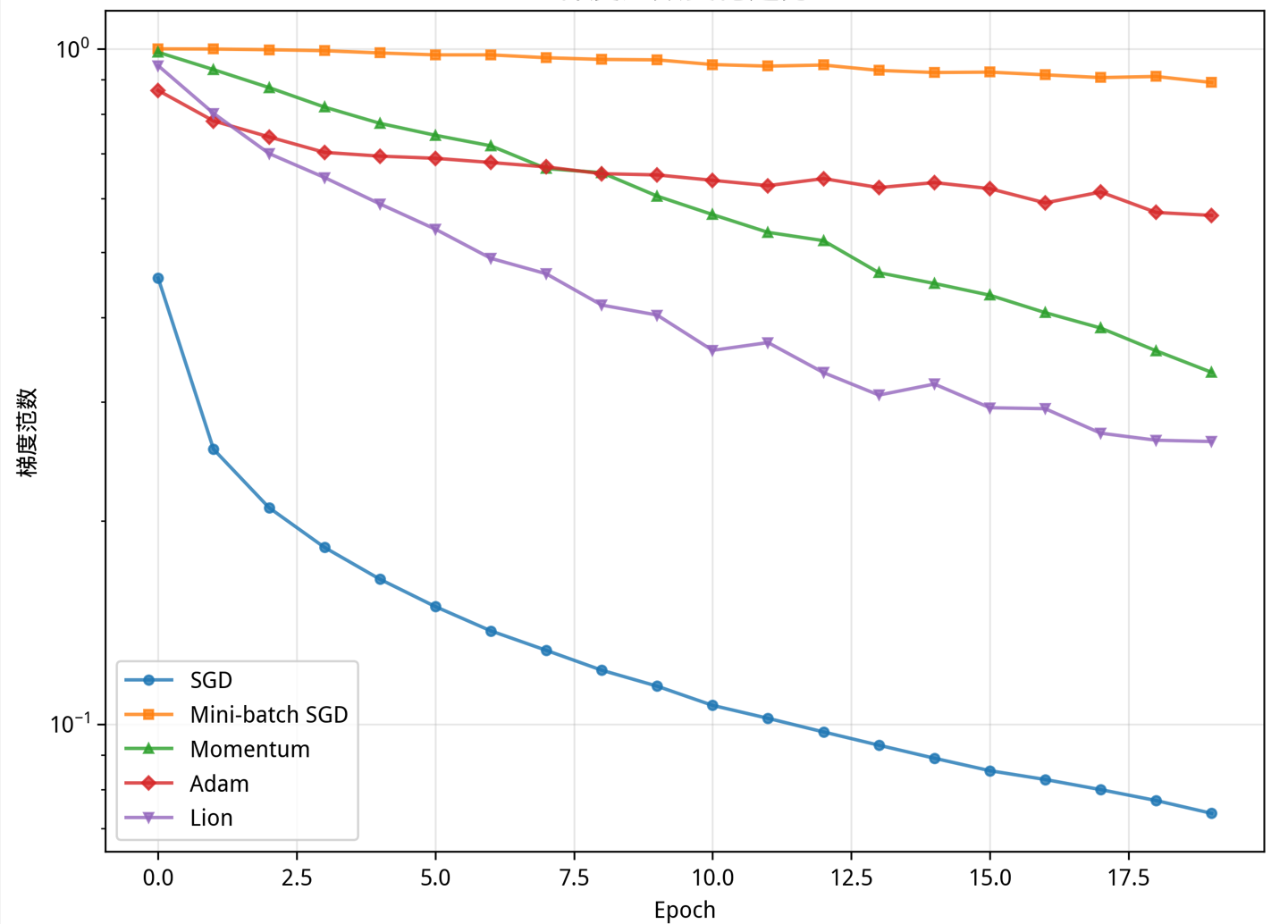}
\caption{Trends of Gradient Norms of Different Algorithms on MNIST}
\label{fig_5}
\end{figure}

\begin{figure}[htbp]
\centering
\includegraphics[width=2.5in]{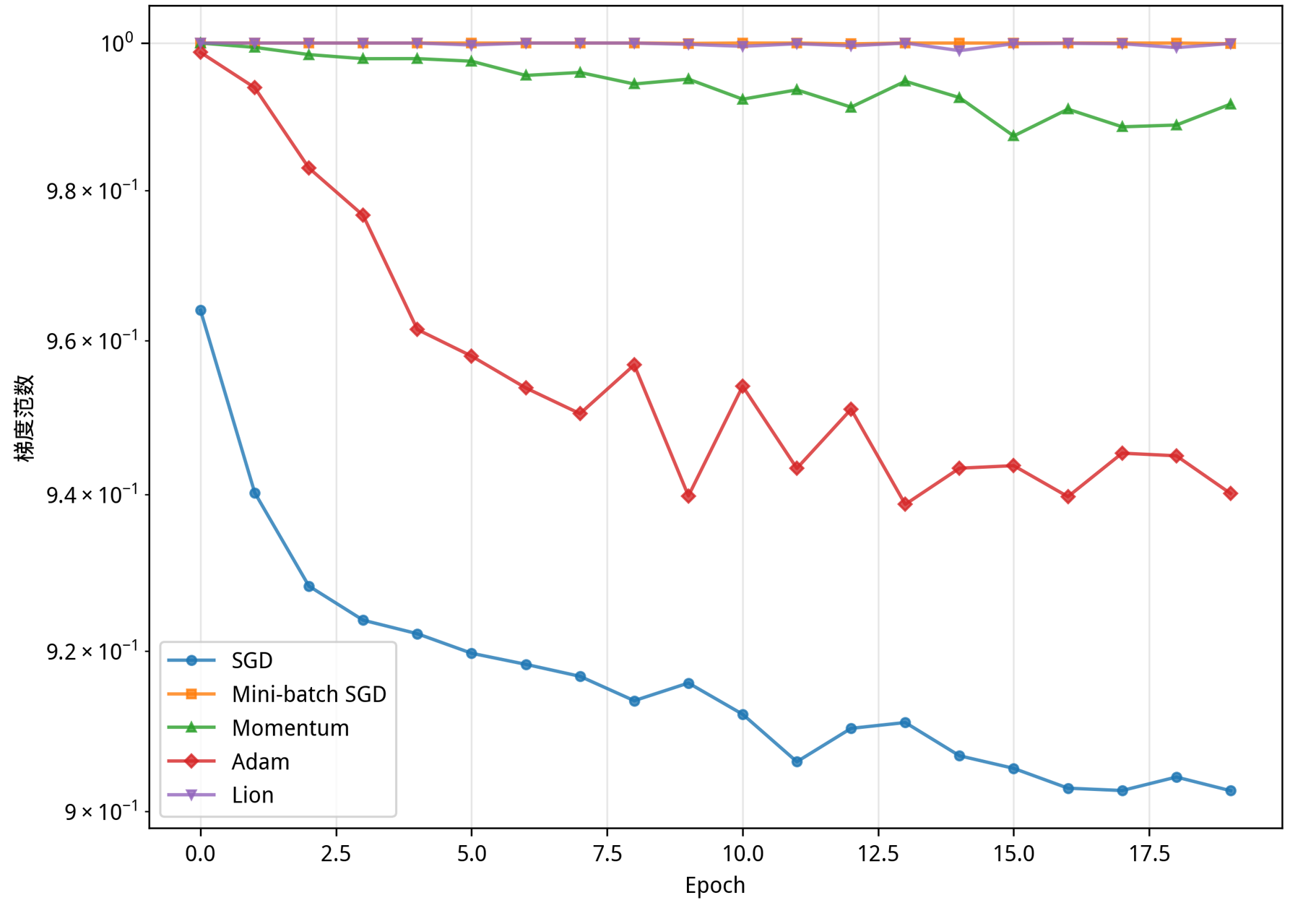}
\caption{Trends of Gradient Norms of Different Algorithms on California Housing}
\label{fig_6}
\end{figure}

The trends of gradient L2 norms further reveal the behavioral differences between optimizers. The gradient norm of SGD decreases significantly during training and shows a relatively regular convergence trend, indicating that it can continuously stabilize gradients and gradually approach the optimal region under long-term training. The gradient norm of Mini-batch SGD remains at a high level with significant fluctuations, meaning its update direction is subject to excessive random perturbations—this phenomenon is highly consistent with the oscillations in its loss curve.

Momentum effectively smooths gradients through momentum accumulation, with the gradient norm decreasing slowly but more consistently, especially on MNIST. This is one of the reasons for its fastest convergence. The gradient norm curve of Adam exhibits considerable fluctuations; its adaptive learning rate mechanism tends to amplify gradient noise without additional hyperparameter tuning, causing frequent changes in update step sizes between different epochs, which leads to subsequent oscillations in the loss curve.

The gradient norm of Lion maintains a low-amplitude decrease with the smallest overall fluctuations in both tasks, indicating that the two-step update rule based on gradient signs is highly effective in suppressing gradient noise and insensitive to the distribution of gradient magnitudes. This stability is not only reflected in the loss curves but also ultimately in the better generalization of test metrics.

\subsection{Quantitative Results and Discussion}

\begin{figure}[htbp]
\centering
\includegraphics[width=2.5in]{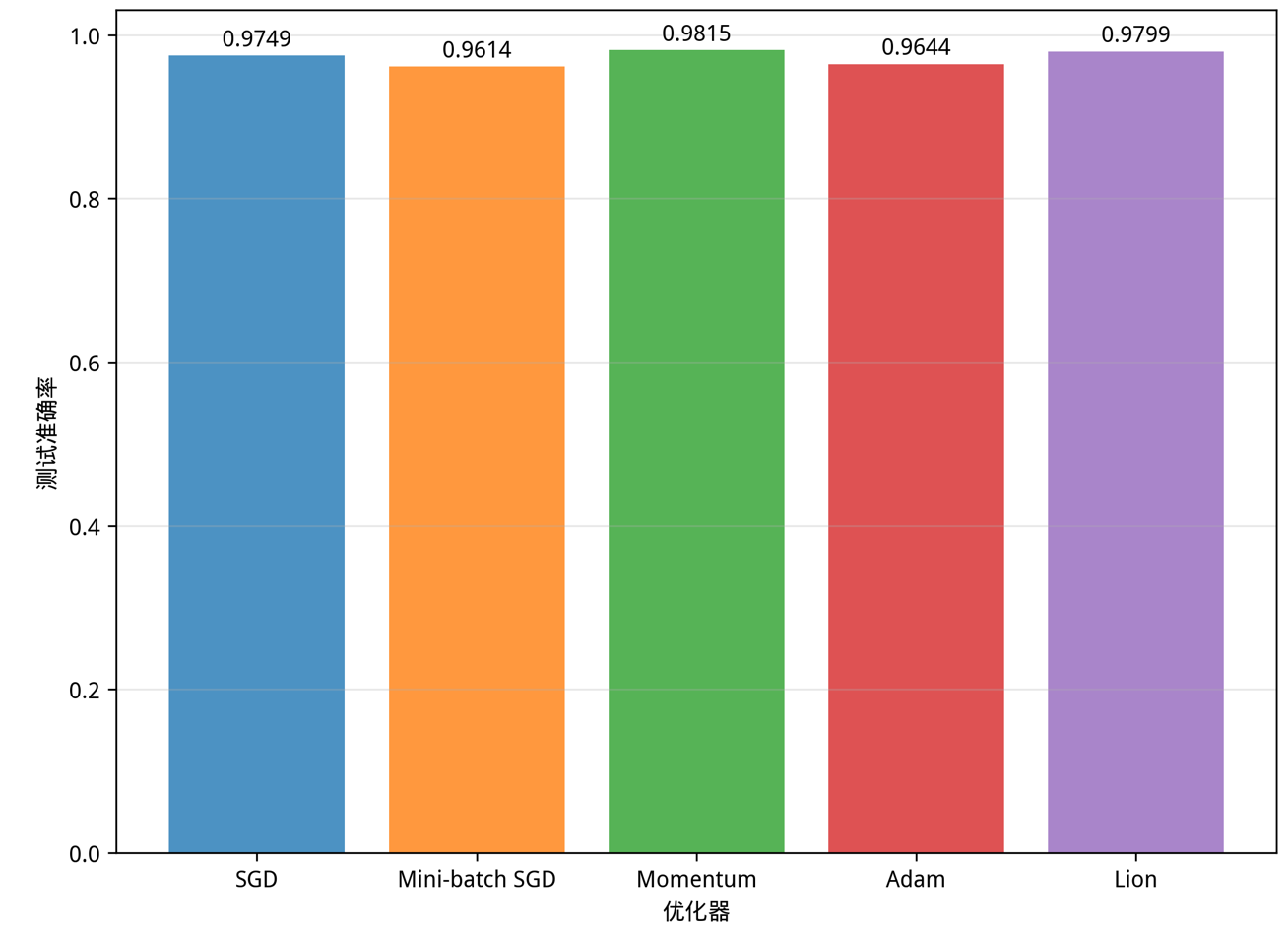}
\caption{Test Accuracy of Different Algorithms on MNIST}
\label{fig_7}
\end{figure}

\begin{figure}[htbp]
\centering
\includegraphics[width=2.5in]{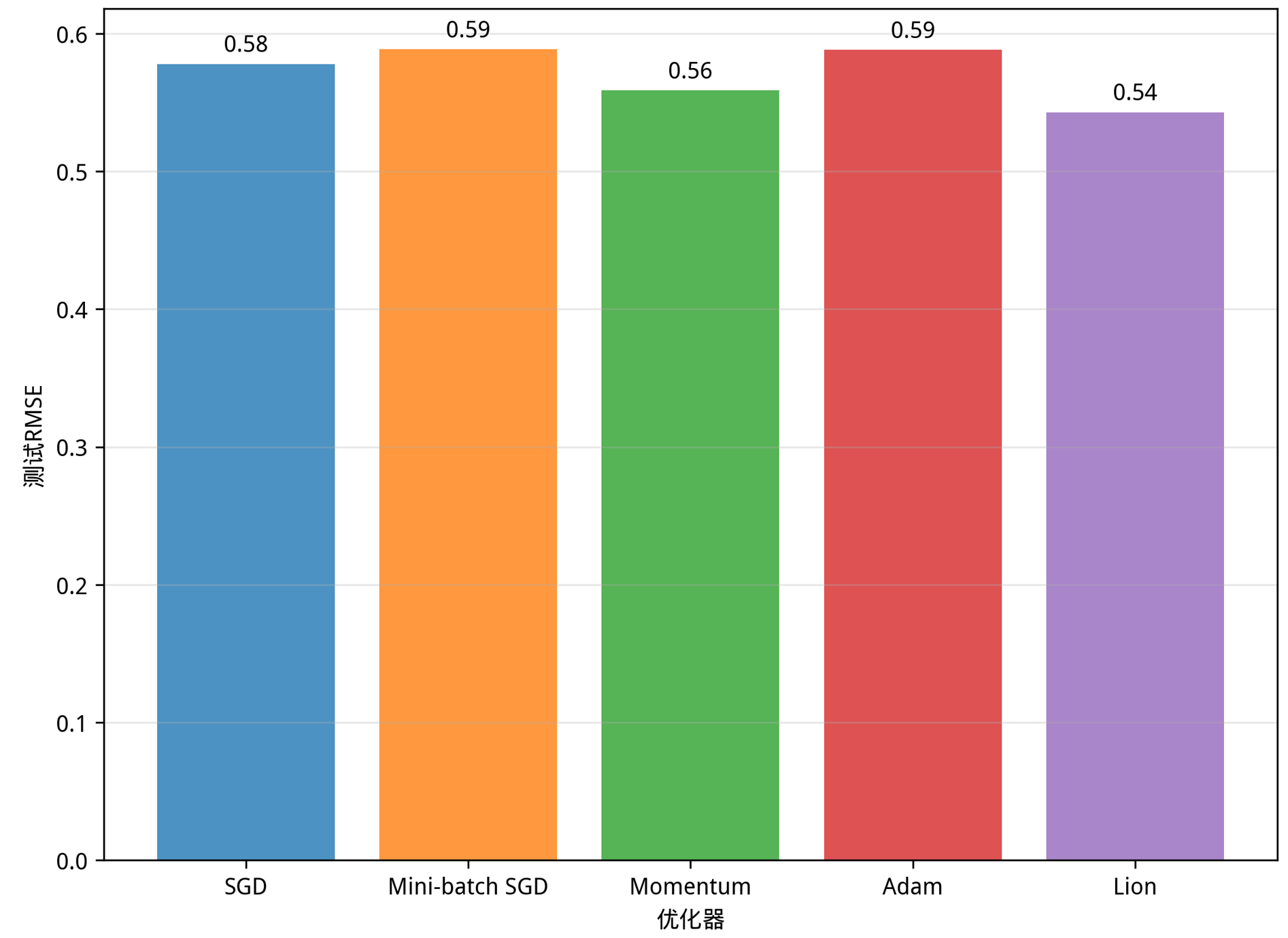}
\caption{RMSE of Different Algorithms on California Housing Test Set}
\label{fig_8}
\end{figure}

In the final test accuracy on MNIST, Momentum achieves the highest performance of 0.9815, followed by Lion with 0.9799, while SGD maintains a high level of 0.9749. Adam and Mini-batch SGD are significantly behind, with 0.9644 and 0.9614 respectively. Overall, Momentum and Lion have obvious advantages in classification tasks, and SGD, although traditional, still maintains high reliability.

In the California Housing regression task, Lion achieves the smallest RMSE (0.54), demonstrating the best generalization ability, followed by Momentum (0.56) and SGD with stable performance (0.58). Adam and Mini-batch SGD both have RMSE values around 0.59, making them the two worst-performing methods. The consistent results of the two tasks show that Lion is a robust and superior optimizer across different data structures and task types, while Momentum is particularly effective in data with clear structures and relatively controllable gradient noise.

\section{Algorithm Summary and Application Recommendations}

Synthesizing the experimental results of the two tasks, it can be seen that the performance of each optimizer is highly consistent with its design philosophy. Lion's design, which relies on gradient signs rather than magnitudes, enables it to maintain a stable and rapid descent trend on noisy data, achieving excellent performance in both tasks and making it suitable for use as a general-purpose optimizer. Momentum demonstrates the strongest competitiveness in tasks with clear structures and relatively controllable gradient noise, especially in the MNIST classification task, making it a highly practical choice among traditional optimization methods.

Although SGD has a slower convergence speed than other methods, its stability, generalization ability, and simplicity make it still an important tool for constructing baselines and analyzing model behavior. In contrast, the performance of Mini-batch SGD is highly dependent on the selection of batch size; without matching momentum or appropriate hyperparameters, it is prone to being affected by noise and resulting in unstable convergence. Although Adam is often preferred in many complex tasks, it performs poorly in this experiment, further indicating that its default hyperparameters are highly sensitive to different tasks. Usually, more sufficient hyperparameter tuning or improved versions such as AdamW are required to achieve the desired results.

In summary, after comprehensively considering the generalization performance, training stability, and task adaptability of the optimizers, Lion and Momentum can be regarded as the two most advantageous choices under the experimental conditions of this study, while SGD remains a robust and reliable baseline method. The performance of Adam and Mini-batch SGD reminds us that the use of optimization methods often depends on task characteristics and hyperparameter settings, and cannot be selected solely based on experience or default configurations. Future work can further extend the verification of optimizers in more complex network structures, large-scale datasets, and different types of tasks to obtain more generalizable conclusions.

\section{Conclusion}

This paper focuses on a systematic investigation of five representative gradient descent optimization algorithms in deep learning (SGD, Mini-batch SGD, Momentum, Adam, and Lion), aiming to provide a standardized reference for algorithm selection and hyperparameter tuning in academic research and engineering practice. First, from a theoretical perspective, it analyzes the historical background, core principles, mathematical properties, and engineering practice configuration guidelines of each algorithm: SGD, as a foundational algorithm, possesses both computational and statistical advantages; Mini-batch SGD balances efficiency and stability, becoming the mainstream in engineering; the Momentum method accelerates convergence through gradient accumulation; Adam integrates momentum with adaptive learning rates to achieve rapid convergence; and Lion achieves memory efficiency and adaptability to large-scale tasks via symbolic momentum updates. Subsequently, experiments are conducted based on the MNIST classification dataset and the California Housing regression dataset, with comparative analysis from dimensions such as convergence speed, gradient characteristics, and generalization performance. The results show that Lion exhibits robust and superior performance on noisy data and large-scale tasks, the Momentum method demonstrates prominent competitiveness on low-noise and highly structured data, SGD has strong stability despite slow convergence, while the performance of Mini-batch SGD and Adam is significantly influenced by hyperparameters and task characteristics. Finally, the applicable scenarios of each algorithm are summarized, indicating that Lion and the Momentum method are the preferred choices under the experimental conditions, and SGD serves as a reliable baseline. It also emphasizes that the use of optimization algorithms needs to be combined with task characteristics and hyperparameter tuning, and future work can further verify the generalizability of the algorithms in more complex scenarios.

\end{document}